\documentclass[11pt]{article}

\usepackage[preprint]{acl}

\usepackage{times}
\usepackage{latexsym}
\usepackage[T1]{fontenc}
\usepackage[utf8]{inputenc}
\usepackage{microtype}
\usepackage{inconsolata}
\usepackage{graphicx}
\usepackage{booktabs}
\usepackage{url}
\usepackage{amsmath}
\usepackage{amssymb}

\title{Stoicheia: Character-Level Masked Diffusion for Ancient Greek\\
       Textual Restoration, Parsing, and Metrical Scansion}

\author{Eric Cullhed \\
  Department of Linguistics and Philology \\
  Uppsala University \\
  Sweden \\
  \texttt{eric.cullhed@lingfil.uu.se} \\\And
  Albin Thörn Cleland \\
  Centre for Languages and Literature \\
  Lund University \\
  Sweden \\
  \texttt{albin.thorn\_cleland@klass.lu.se} \\}

\newcommand{\hforgurl}{\url{https://huggingface.co/collections/Ericu950/stoicheia-6a6fbf9800c82d93020a7ceb}}
\newcommand{\codeurl}{\url{https://github.com/ericu9500/stoicheia}}
\newcommand{\artifactfootnote}{Models and data are collected at \url{https://huggingface.co/collections/Ericu950/stoicheia-6a6fbf9800c82d93020a7ceb}, code at \url{https://github.com/ericu9500/stoicheia}, which also opens the demonstration notebook in Colab.}

\begin{document}
\maketitle

\begin{abstract}
We introduce Stoicheia, a 405M-parameter character-level masked-diffusion encoder for Ancient Greek whose input factors into five aligned, independently maskable planes: letters, word and sentence boundaries, diacritics, capitalization, and punctuation. A single backbone can therefore restore lacunae, re-segment, accentuate, and punctuate unspaced text without task-specific retokenization. We pretrain it on an open, revision-pinned corpus of $\sim$361M words and release eleven checkpoints: ten rotated, decontaminated folds -- guaranteeing that, for any given literary passage, at least one released model has never seen its text -- and one with no exposure to documentary texts. Three experiments -- reconstruction of damaged inscriptions and papyri, morphosyntactic tagging and dependency parsing, and macronization with metrical scansion -- each carry a matched random-initialization control, isolating what character-level diffusion pretraining contributes: 5.6 CER points on inscription reconstruction, 12.9 LAS on parsing, and 6.0 points of balanced accuracy on macronization. On Ithaca's own test split, with identical frozen samples and strict scoring, Stoicheia reduces character error relative to both prior state-of-the-art systems, from 24.6 (Ithaca) and 23.5 (its 2025 Aeneas-framework successor) to 15.5, and raises top-1 accuracy from 63.0 and 64.0 to 74.5.
\end{abstract}

\section{Introduction}
\label{sec:intro}

Neural encoders have become standard tools in computational research on historical languages, including Ancient Greek \citep{sommerschield-etal-2023-survey}. Pretrained models such as Ancient Greek BERT \citep{singh-etal-2021-pilot}, GreBERTa and its siblings \citep{riemenschneider-frank-2023-exploring}, and Logion \citep{cowen-breen-etal-2023-logion} support morphosyntactic tagging, lemmatization, and textual restoration. These models adapt an efficient general-purpose NLP paradigm to Ancient Greek, but it is not obvious that this approach is optimal either for the data -- a finite corpus of often fragmentary, heterogeneous texts -- or for the use cases that philological research involves. Purpose-built architectures show the promise of designing for the task instead: dedicated neural systems have been developed to restore, date, and attribute damaged Greek inscriptions \citep{assael2019pythia,assael2022ithaca} and, most recently, Latin ones \citep{assael2025aeneas}. Domain-specific adaptation of general-purpose large language models has also delivered strong results: an instruction-tuned Llama model achieved slightly better epigraphic restoration than those dedicated systems, including on previously unedited inscriptions and papyri that demonstrably could not have appeared in its training data \citep{cullhed2024instruct}. That guarantee, however, could be given only because the test material was new. For the vast majority of the corpus -- texts long since edited, digitized, and circulated -- no such assurance is possible while training data remains undisclosed \citep{balloccu-etal-2024-leak}. The path pioneered by Pythia, Ithaca, and Aeneas -- controlled, purpose-built models whose training data and held-out structure are explicit -- is, we argue, the more sustainable one for digital philology, and it should be extended beyond inscriptions to all Greek and Latin texts. What must such a model provide? Attempts to adapt the existing encoders to this role surface at least four desiderata.

\paragraph{Closed and undocumented data.}
The highest-quality machine-readable corpus of Ancient Greek, the \textit{Thesaurus Linguae Graecae}, is distributed under a license that prohibits redistribution.\footnote{\url{https://stephanus.tlg.uci.edu/copyright.php}} Open alternatives exist and have recently been gathered in the 40M+-token \textit{Opera Graeca Adnotata} \citep{celano2024oga}. But the largest pretraining efforts to date have not made their corpora available. \citet{riemenschneider-frank-2023-exploring} assembled over 100M words, mixing canonical editions with cleaned Internet Archive OCR, and released their models but not a redistributable copy of the corpus \citep[cf.][]{krahn-etal-2023-sentence}; Logion's training data incorporates the license-restricted TLG by special permission and likewise cannot be shared \citep{cowen-breen-etal-2023-logion,graziosi-etal-2023-tapa}; and the same holds for the pretraining data behind the general-purpose LLMs adapted by \citet{cullhed2024instruct}. The result is that the field's strongest pretrained models cannot be reproduced.

\paragraph{Subword tokenization.}
Subword vocabularies are computationally efficient, and pretrained Ancient Greek language models have adopted them for that reason. But many philological tasks require labeling or prediction at positions that no preset token boundary respects: a lacuna is measured in letters and a metrical quantity belongs to a syllable. Character-level resolution is indispensable. The DeepMind systems drew this consequence at design time: Pythia, Ithaca, and Aeneas all operate directly on characters, precisely so that damage of arbitrary extent has a representation at the input \citep{assael2019pythia,assael2022ithaca,assael2025aeneas}. Token-free pretraining is established in general NLP \citep{clark-etal-2022-canine,xue-etal-2022-byt5}, and character-level baselines with randomly initialized embeddings have recently proven competitive for Ancient Greek morphology \citep{celano2025parser}. We lack a character-level encoder for Ancient Greek that brings the benefits of large-scale pretraining to the input representation philological tasks typically require.

\paragraph{Editorial layering.}
A digitized edition is not a facsimile of its source. Ancient copies were written without accents and largely without word division or punctuation; the medieval manuscripts that transmit most literary texts do carry accents and divisions, but these are themselves products of transmission, added and reinterpreted by generations of copyists. Modern editions further supply punctuation, capitalization, and restorations of lost or corrupted text. A model that ingests accented, pre-divided words as ground truth therefore hard-codes the tradition's interpretation of the sequences of signs, and cannot easily be used to question it.

\paragraph{Memorization.}
A model trained on everything is, for the philologist, a model usable on almost nothing. Language models demonstrably memorize training data, increasingly so with scale \citep{carlini2023quantifying}. Ithaca \citep{assael2022ithaca} may legitimately be used on unedited inscriptions, or on its own held-out validation and test sets (inscriptions whose PHI identifiers end in 3 and 4) -- but for the remaining $\sim$80\% of the corpus, any restoration it proposes as an alternative to an earlier scholarly reconstruction is inescapably conditioned by it. This holds a fortiori for general pretrained models covering the whole surviving corpus of Greek texts. Much of the value of computational methods for philology lies not in editing new texts but in interrogating the tradition itself \citep{cowen-breen-etal-2023-logion}; that requires models whose ignorance of a given passage is guaranteed.

\smallskip

\noindent Stoicheia is our attempt to meet these demands. We present:
\textbf{(1)} an open, revision-pinned $\sim$361M-word pretraining corpus combining openly licensed editions with re-OCR'd Internet Archive material in documented quality tiers, plus a synthetic corpus;
\textbf{(2)} a character-level masked-diffusion encoder whose input separates letters, diacritics, word and sentence boundaries, capitalization, and punctuation into five aligned planes, so that one backbone handles edited text or unaccented \textit{scriptio continua};
\textbf{(3)} eleven released checkpoints -- ten with rotated, decontaminated 80/10/10 splits of the literary corpus, so that at least one has provably never seen the text of any given passage, and one with zero documentary exposure as a leak-proof base for epigraphic and papyrological work;
\textbf{(4)} three fine-tuning experiments -- reconstruction of damaged inscriptions and papyri, morphosyntactic tagging and dependency parsing, and macronization with metrical scansion -- each quantifying what pretraining buys through a matched random-initialization control, both arms trained to convergence;
and \textbf{(5)} same-harness evaluation against the strongest available baselines: on reconstruction, against Ithaca and its 2025 Aeneas-framework successor \citep{assael2025aeneas} on their own test split with identical frozen samples, and -- together with an 8B instruction-tuned Llama \citep{cullhed2024instruct} -- on documents edited only after every compared system's training-data cutoff; on parsing, against learning-rate-tuned Ancient Greek and multilingual subword encoders under the same recipe; and on macronization, against the character transformer of \citet{thorncleland2026vowel}, trained on that paper's rule-based silver.

\section{Data}
\label{sec:data}

Answering the first desideratum of \S\ref{sec:intro}, our pretraining corpora are openly released: a corpus of genuine Ancient Greek ($\sim$361M words across nine sources) and a synthetic augmentation corpus machine-translated from Latin.\footnote{\artifactfootnote}

Two public-domain page-image collections supply the bulk: Internet Archive volumes ($\sim$232M words) and PleIAs Greek-PD ($\sim$89M). Born-digital editions contribute the rest, each under its own license: \textit{Opera Graeca Adnotata} \citep[$\sim$34M;][]{celano2024oga}, papyri.info \citep[$\sim$4M;][]{papyri-info}, the calfa OCR of the \textit{Patrologia Graeca} \citep[$\sim$3M;][]{patrologia-graeca}, and $\sim$0.3M from \textit{Corpus Corporum}, Project Gutenberg and the SBL New Testament. The Database of Byzantine Book Epigrams \citep{dbbe} was in our pretraining corpus, but its CC-BY-NC-SA terms bar it from a share-alike release; the code ships a script that fetches it. Rather than reusing the Archive's existing text layer, we re-OCR'd every Greek page with a vision--language model fine-tuned for polytonic Greek \citep{qwen3vl-ocr-paper}, roughly quadrupling the usable yield. All text then passed through a normalization and dictionary-based correction stage built on an Ancient Greek Hunspell dictionary \citep{hunspell-grc} expanded through iterative manual review. Each document's resulting cleanliness score assigned it to a \textbf{pristine} tier ($\sim$90M words, released untouched) or a \textbf{repaired} tier ($\sim$271M words, reconstructed by an instruction-tuned LLM \citep[Qwen3.6-27B;][]{qwen36} constrained to correct only well-attested OCR confusions and to discard rather than fabricate).

A separate \textbf{bronze} tier of 676K passages ($\sim$1.4B characters) was machine-translated into Ancient Greek from the Latin holdings of \textit{Corpus Corporum} \citep{corpus-corporum} with the same model.

\section{Model}
\label{sec:model}

\paragraph{Architecture.}
We pretrain \textbf{Stoicheia}, a 405M-parameter character-level masked-diffusion Transformer (\texttt{d\_model}=1024, depth 32, QK-norm) in the spirit of recent discrete diffusion language models \citep{austin2021structured,lou2024discrete,sahoo2024simple}. The input is factored not into subwords but into five aligned character-level planes: base letter identity (a 24-symbol minimal alphabet, with medial, lunate, and final sigma merged), a word/sentence-boundary plane, a diacritic plane (accent $\times$ breathing $\times$ iota subscript $\times$ diaeresis), a capitalization plane, and a punctuation-category plane. Each plane can be masked independently to an explicit ``unknown'' state at inference. Attention is banded rather than dense: three of every four blocks attend within a 256-character window and the fourth attends globally.

\paragraph{Objective.}
Each training sequence is corrupted at a rate $t$ drawn afresh from a clipped Beta distribution, and the model predicts the true content at every corrupted position under the standard $1/t$ diffusion loss reweighting \citep{sahoo2024simple}. What is corrupted is shaped like real damage: contiguous spans that ignore word edges (a break in the stone does not respect them), whole words, partial words anchored to their beginning, middle, or end, scattered single characters, and silent substitution of wrong letters with no mask marker. The most consequential pattern is \emph{elastic} masking: a span of true length $L$ is replaced by $M\ge L$ mask slots, and the target is the $L$ true characters followed by $M{-}L$ copies of an explicit empty symbol $\varnothing$. This teaches variable-length infilling inside a bidirectional encoder. The boundary, diacritic, and punctuation planes are independently set per position to known, unknown, or patchy states.

\paragraph{Pretraining regime.}
Training is dev-driven: warmup, then a constant peak learning rate held until the held-out masked-bits-per-character metric stalls, then a cosine decay. Each released checkpoint is the best held-out bits-per-character step rather than the last, so the eleven differ in length (78K--122K steps). During this anneal phase the bronze (synthetic) tier is downweighted first, then the silver (rule-labeled) tier, ending with gold data only. Staged annealing converged to a better final dev metric than a hard-switch ablation at matched compute.

\paragraph{The fleet.}
We release \textbf{eleven} pretrained checkpoints. Ten of them, \path{Stoicheia_fold_0} through \path{Stoicheia_fold_9}, instantiate a rotated 80/10/10 train/dev/test split of the literary corpus. Where canonical identifiers exist, work identity is given -- OGA records carry TLG work identifiers -- but most of the corpus is not so labeled: the re-OCR'd Internet Archive volumes contain further alternative editions of the same works under no shared identifier, and many texts are quoted or paraphrased at length inside later ones, so a naive document-level split leaks nominally held-out text back into training. A 13-stage pipeline therefore clusters records into editions of the same work -- by identifier where one exists, by duplicate-aware content matching elsewhere -- assigns every cluster to one of ten zones, and excises from each fold's training data every sentence colliding with its held-out zones, matching exact and reordered duplicates, word 5-grams, and document-level near-duplicates. The exclusion is exact at the fold level: at least one released model has provably never seen the text of any given passage of the open corpus, so reconstruction experiments on held-out text are possible without circularity.

The eleventh checkpoint (\path{Stoicheia_doc_clean}) extends the same discipline to documentary text: every inscription and papyrus record is excluded, along with every passage from the other sources that a contamination screen -- exact, bag-of-words, and 8-gram matching plus whole-record MinHash-LSH and an explicit blocklist -- flags as quoting or closely paraphrasing one. The result is a decontamination guarantee for epigraphic and papyrological fine-tuning. All eleven models converge to a comparable final dev bits-per-character (fold 0 $\approx$0.355, fold 9 $\approx$0.365, doc-clean $\approx$0.234/0.309 on pristine/repaired reference slices), each after roughly forty-two hours of training on 128 NVIDIA GH200 GPUs.
\section{Experimental design}
\label{sec:design}

In each experiment, one pretrained backbone (\S\ref{sec:model}) is equipped with task-specific heads and fine-tuned under the same recipe. Every experiment also carries a matched control with the pretrained weights replaced by random ones. The gap between the arms isolates the contribution of pretraining. Both arms are trained to convergence rather than to a shared step budget: a random-init arm stopped when the pretrained arm stops is typically still improving, and truncation would inflate the apparent value of pretraining.

Significance is assessed with paired tests throughout. Across matched runs -- fine-tuning seeds or rotation folds -- we use exact sign-flip permutation tests, with paired $t$-tests and effect sizes alongside. Where two systems read the identical samples, their per-sample outcomes are paired, not independent, so head-to-head accuracies are compared with McNemar's exact test on the discordant pairs. The permutation test assumes nothing about the distribution of the differences, but its precision is bounded by the number of pairs: with $n$ pairs there are only $2^n$ sign assignments, so even when all differences agree in direction the two-sided $p$ can go no lower than $2/2^n$ (0.0625 at $n{=}5$, 0.031 at $n{=}6$, 0.0078 at $n{=}8$, 0.0039 at $n{=}9$, 0.002 at $n{=}10$). We flag such floor values where they occur; they mark the resolution limit of the test rather than a weak effect. External baselines for all three experiments are evaluated in our own harness, on the same evaluation data as our model, with fine-tuned baselines run at their own tuned learning rates. Every artifact is public: twenty-four checkpoints and five datasets at \hforgurl; code, split pipeline, per-fold manifests, frozen sample files and a Colab notebook that reproduces the demonstrations at \codeurl.

\section{Documentary reconstruction}
\label{sec:restoration}

The first experiment concerns the reconstruction of damaged inscriptions and papyri.

\paragraph{Method.}
We fine-tune the documentary-clean backbone (\S\ref{sec:model}) on inscriptions and papyri, using the rotation described below. At inference, letters in a lacuna are decoded with the word-boundary plane held unknown throughout the gap; context is the whole document wherever it fits the model's window. Two fine-tuning revisions are reported. In v1, the mixture consists of whole inscriptions and whole papyri (weight 1.0 each) with a 4,096-character context. In v2, the mixture expands to six tiers -- the same whole documents, synthetic inscriptions with word order varied and synonyms swapped by GPT-4o \citep{openai2024gpt4o,cullhed2024instruct}, segments split at lacunae in both domains, and a 60K-record literary background sample against catastrophic forgetting -- at an 8,192-character context, with weights as character-share targets. Learning rate and schedule are shared.

\paragraph{Evaluation and metrics.}
The context is the entire document with its real surviving lacunae left in place as unknowns, and an artificial gap of length $L$ must be filled, where spaces count as characters toward $L$ and the word division inside the gap must itself be predicted. Gaps run from $L{=}1$ to $L{=}10$ characters, in equal numbers at each length, following the range reported by \citet{assael2022ithaca}. Character error rate (CER) is the Levenshtein distance between the predicted and the gold string divided by the gold length. Two accuracy metrics accompany it: top-1 is the proportion of gaps whose highest-scoring hypothesis is precisely the gold string, top-20 the proportion whose gold string appears anywhere among the model's twenty best hypotheses. Those hypotheses come from a beam search of width 20 over the gap, deduplicated and ranked by length-normalized log-probability, so top-20 is the candidate list an editor would be shown; every system compared here decodes at that same width.

\paragraph{Ten-fold rotation.}
Following \citet{assael2022ithaca}, we split by the last digit of each document's unique identifier: the Packard Humanities Institute (PHI) number for inscriptions, the Trismegistos (TM) number for papyri. This is arbitrary with respect to a document's content, date, and provenance. Where they fix a single split, we rotate through all ten: ten fine-tuned models per revision, each blind to a different tenth, so every document has a model that provably never saw it in fine-tuning. Table~\ref{tab:restoration-tenfold} reports both revisions and the v1 random-init control on frozen files of 1{,}000 samples per fold and domain (100 per gap length). Across the full rotation, v2 improves on v1 everywhere, reaching 16.10 CER and 73.3 top-1 on inscriptions and 10.01 CER and 81.5 top-1 on papyri.

\begin{table}[t]
\centering
\footnotesize
\setlength{\tabcolsep}{2.5pt}
\begin{tabular}{llcccc}
\toprule
\textbf{Arm} & \textbf{Domain} & \textbf{n} & \textbf{CER}$\downarrow$ & \textbf{Top-1}$\uparrow$ & \textbf{Top-20}$\uparrow$ \\
\midrule
v1 & inscr. & 10 & 17.36 $\pm$ 1.54 & 71.59 & 79.79 \\
v1 & pap. & 10 & 10.74 $\pm$ 0.88 & 80.15 & 87.15 \\
v2 & inscr. & 10 & \textbf{16.10} $\pm$ 1.25 & \textbf{73.33} & \textbf{80.93} \\
v2 & pap. & 10 & \textbf{10.01} $\pm$ 1.05 & \textbf{81.53} & \textbf{88.90} \\
\midrule
v1 rand-init & inscr. & 10 & 23.00 $\pm$ 2.01 & 63.54 & 73.57 \\
v1 rand-init & pap. & 10 & 14.16 $\pm$ 1.00 & 74.91 & 83.31 \\
\bottomrule
\end{tabular}
\caption{Ten-fold rotation (\%, mean $\pm$ sd over folds; per-digit frozen files of 1{,}000 samples per fold and domain). Each fold holds out one PHI/TM digit for test, one for dev; $n$ is not comparable to Table~\ref{tab:restoration-strict}.}
\label{tab:restoration-tenfold}
\end{table}

A seed-repetition check on one fold (the v2 recipe at three seeds) puts seed-to-seed spread at 0.37 CER in both domains, well below the fold-to-fold spread. Replacing the pretrained backbone with the matched random-init control of \S\ref{sec:design}, and changing nothing else in the v1 recipe, costs 5.64 CER points on inscriptions and 3.42 on papyri across the full rotation (Table~\ref{tab:restoration-tenfold}). All ten per-fold differences run in the same direction in both domains ($p{=}0.00195$, the $n{=}10$ floor), so the effect does not rest on a few favorable splits. We release all ten v2 checkpoints, so any document can be interrogated by a model that never saw it, without the circularity of applying a fixed-split model outside its test set.

\paragraph{Comparison with Ithaca and Aeneas.}
Ithaca's published split -- also used by Aeneas \citep{assael2025aeneas} -- fixes the test set to the inscriptions whose PHI identifier ends in 3 \citep{assael2022ithaca}. The Greek model we evaluate ships with that release's software \citep{predictingthepast}, not with the paper, and we run its own inference code. The digit-3 model of our rotation can therefore be set against them: it never saw those documents in fine-tuning, and its backbone never saw a documentary text. All three read the same frozen file of 3{,}000 samples, 300 at each gap length $L{=}1$--$10$, at beam width 20 (Table~\ref{tab:restoration-strict}).

\begin{table}[t]
\centering
\footnotesize
\setlength{\tabcolsep}{3pt}
\begin{tabular}{lccc}
\toprule
\textbf{System} & \textbf{CER}$\downarrow$ & \textbf{Top-1}$\uparrow$ & \textbf{Top-20}$\uparrow$ \\
\midrule
Ithaca & 24.6 & 63.0 & 74.9 \\
Aeneas (2025) & 23.5 & 64.0 & 75.5 \\
Stoicheia v1 & 17.0 & 72.0 & 80.8 \\
Stoicheia v2 & \textbf{15.5} & \textbf{74.5} & \textbf{81.7} \\
\bottomrule
\end{tabular}
\caption{Head-to-head on Ithaca's test split (\%): the same frozen 3{,}000 samples (300 per gap length, $L{=}1$--$10$) read by all systems, beam 20 throughout. All figures are recomputed in our harness, so they are not comparable with those printed in the Ithaca and Aeneas papers.}
\label{tab:restoration-strict}
\end{table}

Stoicheia v2 reduces CER by 37\% relative to Ithaca and 34\% relative to Aeneas, and raises top-1 accuracy from 63.0 and 64.0 to 74.5. Because all three systems read the same frozen file, the comparison is paired and we test it with McNemar's exact test on the discordant samples (\S\ref{sec:design}). Against Ithaca, 439 samples are recovered by us alone and 93 by Ithaca alone ($p{=}9.7\times10^{-55}$); against Aeneas, 415 against 100 ($p{=}1.3\times10^{-46}$); the top-20 margins are likewise significant ($p\leq3.0\times10^{-21}$). Aeneas's own gain over Ithaca on this protocol is not significant (197 against 166 discordant samples, $p{=}0.12$). One limitation is inherent: Ithaca's published split is fixed, so this head-to-head can only ever be run on the one fold.

\paragraph{Comparison on recently edited documents.}
The release accompanying \citet{cullhed2024instruct}, an instruction-tuned Llama 3.1 8B for this task, offers two sets of documents edited \emph{after} every compared system's training data was collected. It ships Ithaca's and the tuned Llama's per-record predictions; we add Stoicheia and, by running the 2025 release's own inference code at beam width 20, Aeneas. Because the release scores character error with a \texttt{difflib} ratio, we verified a port of its scorer against the released artifacts (zero per-record difference) and rescored every system with the Levenshtein CER used throughout, under the release's own normalization. Its documents, gap construction, and normalization differ from our protocol, so these numbers stand apart from the tables above; within them, however, every system is scored identically on identical samples.

\begin{table}[t]
\centering
\small
\setlength{\tabcolsep}{3pt}
\begin{tabular}{lcccc}
\toprule
\textbf{System} & \textbf{Params} & \textbf{CER}$\downarrow$ & \textbf{Top-1}$\uparrow$ & \textbf{Top-20}$\uparrow$ \\
\midrule
Ithaca & -- & 36.50 & 48.41 & 60.67 \\
Llama 3.1 8B & 8B & 34.18 & 51.28 & 68.62 \\
Aeneas (2025) & -- & 33.96 & 52.98 & 62.93 \\
\midrule
Stoicheia v1, rand-init & 405M & 38.83 & 44.64 & 58.33 \\
Stoicheia v1 & 405M & 24.02 & 62.59 & \textbf{73.46} \\
\textbf{Stoicheia v2} & 405M & \textbf{23.45} & \textbf{64.41} & 73.34 \\
\bottomrule
\end{tabular}
\caption{Recently edited inscriptions (4{,}111 gaps in six documents; \%): documents edited after every system's training data was collected, identical samples per row \citep{cullhed2024instruct}, Levenshtein CER under the release's normalization. The lower block is ours: v2, v1, and v1's random-init control. Not comparable to the protocol above.}
\label{tab:recent-inscriptions}
\end{table}

\begin{table}[t]
\centering
\small
\setlength{\tabcolsep}{3pt}
\begin{tabular}{lcccc}
\toprule
\textbf{System} & \textbf{Params} & \textbf{CER}$\downarrow$ & \textbf{Top-1}$\uparrow$ & \textbf{Top-20}$\uparrow$ \\
\midrule
Llama 3.1 8B & 8B & 35.65 & 50.37 & \textbf{70.51} \\
\midrule
Stoicheia v1, rand-init & 405M & 37.10 & 48.14 & 61.11 \\
Stoicheia v1 & 405M & \textbf{30.93} & 56.24 & 67.10 \\
\textbf{Stoicheia v2} & 405M & 31.22 & \textbf{56.67} & 68.40 \\
\bottomrule
\end{tabular}
\caption{Recently published papyri (1{,}620 gaps in two Uppsala papyri; the released Llama predictions cover 1{,}614). Ithaca and Aeneas are inscriptions-trained and therefore absent. Same scoring as Table~\ref{tab:recent-inscriptions}.}
\label{tab:recent-papyri}
\end{table}

Stoicheia leads both sets on CER and top-1: relative error reductions of 31--36\% against the three inscription baselines and 12.4\% against Llama on papyri. Every system faces the identical gaps, so the per-gap comparison is paired (McNemar's exact test: $p\leq3.9\times10^{-54}$ on inscriptions, $p{=}1.9\times10^{-5}$ on papyri). The gaps are not independent, however -- they are cut from six recently edited inscriptions and two unedited Uppsala papyri -- so a system that commands one document's formulary accumulates correlated wins, and these figures describe this material rather than documentary Greek at large. Aggregated by document, our top-1 exceeds Ithaca's and Llama's on all six inscriptions (sign test $p{=}0.031$, the $n{=}6$ floor) and Aeneas's on five of six ($p{=}0.22$); on the two papyri we lead Llama on both. On documents no system could have memorized, the matched random-init control of \S\ref{sec:design} falls behind every external system on both sets, while the pretrained arm of the identical recipe beats them all.

\section{Morphosyntactic tagging and parsing}
\label{sec:morphosyntax}

The second experiment applies the shared design of \S\ref{sec:design} to morphosyntactic annotation, where the backbone competes directly with pretrained subword encoders.

\paragraph{Method.}
We fine-tune a joint model for the full annotation stack: factored XPOS heads, an edit-script lemmatizer, and a biaffine dependency parser \citep{dozat2017deep}, all driven by a single shared backbone through an ELMo-style scalar mix of layer representations. Every hyperparameter is held fixed while only the backbone is swapped. Training and evaluation use the 5-fold treebank splits with a shared held-out \texttt{test.conllu} from \citet{celano2025parser}, and each encoder is run over the full 5 folds $\times$ 2 seeds matrix, i.e.\ 10 runs per encoder. Subword encoders are bridged into the same interface: each word's token vectors (aligned via \texttt{word\_ids()}) are pooled into one word vector, exactly as Stoicheia's character positions are. Our entry is the documentary-clean checkpoint (\mbox{Stoicheia\_doc\_clean}). The decontamination of \S\ref{sec:restoration} does not extend here: it withholds documentary, not literary, material, so the backbone has read editions of the works these treebanks annotate -- as have all the subword baselines, which keeps the comparison internally fair. The external baselines are GreBERTa and PhilBERTa \citep{riemenschneider-frank-2023-exploring}, Logion \citep{cowen-breen-etal-2023-logion}, Ancient-Greek-BERT \citep{singh-etal-2021-pilot}, XLM-R base and large \citep{conneau2020unsupervised}, and mBERT \citep{devlin-etal-2019-bert}.

\begin{table}[t]
\centering
\footnotesize
\setlength{\tabcolsep}{3.5pt}
\begin{tabular}{lcccc}
\toprule
\textbf{Encoder} & \textbf{LAS} & \textbf{UAS} & \textbf{XPOS} & \textbf{Lem.} \\
\midrule
Stoicheia (200 ep.) & 83.89 {\scriptsize$\pm$0.17} & 89.17 & 93.31 & 93.89 \\
\textbf{Stoicheia} & \textbf{83.81} {\scriptsize$\pm$\textbf{0.17}} & \textbf{89.13} & \textbf{93.30} & 93.78 \\
PhilBERTa (tuned) & 82.89 {\scriptsize$\pm$0.15} & 88.34 & 92.28 & 93.26 \\
GreBERTa (tuned) & 82.52 {\scriptsize$\pm$0.15} & 88.05 & 92.69 & 93.43 \\
PhilBERTa & 82.25 {\scriptsize$\pm$0.15} & 87.88 & 92.08 & 93.84 \\
GreBERTa & 82.02 {\scriptsize$\pm$0.17} & 87.70 & 92.48 & \textbf{93.91} \\
Logion & 81.46 {\scriptsize$\pm$0.17} & 87.23 & 91.55 & 91.67 \\
XLM-R large & 80.56 {\scriptsize$\pm$0.30} & 86.61 & 91.24 & 93.46 \\
AG-BERT & 78.84 {\scriptsize$\pm$0.20} & 85.13 & 90.11 & 90.67 \\
XLM-R base & 78.79 {\scriptsize$\pm$0.30} & 85.18 & 90.35 & 93.17 \\
Rand-init (200 ep.) & 71.00 {\scriptsize$\pm$0.49} & 78.65 & 88.00 & 92.33 \\
Rand-init & 68.41 {\scriptsize$\pm$0.57} & 76.38 & 86.52 & 91.93 \\
mBERT & 66.96 {\scriptsize$\pm$0.45} & 75.72 & 79.65 & 83.19 \\
\midrule
\multicolumn{5}{l}{\emph{\citet{celano2025parser}:}} \\
tagger--parser & 77.10 & 82.60 & 91.90 & -- \\
GreTa & -- & -- & -- & 91.17 \\
\bottomrule
\end{tabular}
\caption{Tagging and parsing on the 5-fold split of \citet{celano2025parser} over the AGDT, Gorman and Pedalion treebanks (1.26M tokens), shared held-out test split (\%); one recipe throughout, only the backbone swapped. Every row is ten runs (5 folds $\times$ 2 seeds); LAS is mean $\pm$ sd, Lemma a mean (sds 0.14--0.44). ``Tuned'' rows use each baseline's own swept optimum; UPOS is omitted for space (ours 96.94 against 96.41). The final block is the best published on this data, from two systems under their own architectures, so it is not recipe-matched.}
\label{tab:morpho-main}
\end{table}

\paragraph{Results.}
Table~\ref{tab:morpho-main} reports all thirteen configurations, each over the complete 5-fold $\times$ 2-seed matrix. Stoicheia reaches 83.81 LAS, ahead of the strongest external encoder -- learning-rate-tuned PhilBERTa (82.89) -- by +0.92 LAS; paired by fold and seed, this sits at the exact test's floor ($p{=}0.00195$ at $n{=}10$, $d_z{=}6.2$), as does every other paired comparison, from +1.29 over tuned GreBERTa to +16.85 over mBERT. It also exceeds every figure published on this treebank: +6.7 LAS, +6.5 UAS, +1.4 XPOS, and +0.5 UPOS over the best tagger--parser of \citet{celano2025parser}, and +2.6 lemma over the mean reported for GreTa, the best lemmatizer there (+2.4 against its best fold). Among the baselines, Greek-specific pretraining beats family and scale: PhilBERTa, GreBERTa, and Logion all outperform the larger XLM-R large, while mBERT falls below even our converged random-init control. Lemmatization is the one metric where subword baselines edge us in our own harness (shared-recipe GreBERTa 93.91 and PhilBERTa 93.84 against our 93.78); their LR-tuned rows drop to 93.43 and 93.26, because tuning the encoder rate for LAS costs lemma accuracy.

\paragraph{Pretraining ablation.}
On the standard schedule the random-init arm reaches 68.41 LAS while still improving at roughly +1.1 LAS per 10 epochs, so the gap at that point (+15.4) would credit pretraining with unfinished optimization. With both arms run to 200 epochs, the matched comparison is 83.89 against 71.00, a pretraining effect of \textbf{+12.9 LAS} ($p{=}0.002$, the $n{=}10$ floor; $d_z{=}25.3$).

\paragraph{Recipe versus backbone.}
\citet{celano2025parser} report GreBERTa performing poorly on syntax under their own architecture (LAS 53.41), attributing this to the absence of ``a further modeling strategy\ldots such as adjacency matrices or biaffine attention''. Under our recipe the identical checkpoint reaches 82.52 LAS: a swing of roughly 29 points on the same weights.

\section{Macronization and metrical scansion}
\label{sec:meter}

\begin{table}[t]
\centering
\small
\setlength{\tabcolsep}{4pt}
\begin{tabular}{llccc}
\toprule
\textbf{Task} & \textbf{Backbone} & Bal.\ acc. & Acc. & Macro/F1 \\
\midrule
Macronize & pretrained & \textbf{93.31} $\pm$ 0.62 & 94.20 & 91.89 \\
              & rand-init & 87.33 $\pm$ 0.76 & 91.40 & 88.72 \\
\midrule
Scan & pretrained & \textbf{89.74} $\pm$ 0.88 & 88.74 & 87.89 \\
         & rand-init & 87.54 $\pm$ 0.69 & 86.18 & 85.14 \\
\bottomrule
\end{tabular}
\caption{Pretraining ablation (\%, Norma test, $n{=}2{,}660$ positions), six seeds per arm; $\pm$ is sd over seeds. The last column is macro-balanced accuracy for macronization, syllable-boundary F1 for scansion.}
\label{tab:meter}
\end{table}

\begin{table}[t]
\centering
\small
\setlength{\tabcolsep}{4pt}
\begin{tabular}{lccc}
\toprule
\textbf{System} & \textbf{Acc.} & \textbf{Unm.$\to$short} & \textbf{Unmarked} \\
\midrule
All-short baseline & -- & 83.14 & -- \\
Rule-based & 57.57 & 89.46 & 769 \\
Transformer-based & 90.55 & 90.66 & 6 \\
\textbf{Stoicheia} & \textbf{93.37} & \textbf{93.37} & \textbf{0} \\
\bottomrule
\end{tabular}
\caption{Macronization on Norma's 1{,}916 test positions -- a different set and metric from Table~\ref{tab:meter}. External rows from \citet{thorncleland2026vowel}; ours is the mean over the six-seed arm (93.37 $\pm$ 1.87; dev-selected seed 94.52).}
\label{tab:macronizer}
\end{table}

Greek orthography leaves vowel length unmarked: a bare $\alpha$, $\iota$, or $\upsilon$ -- the \emph{dichrona} -- may be long or short depending on lexeme, morphology, sandhi, dialect, and metre \citep{thorncleland2026vowel}. Recovering these lengths (\emph{macronization}) is natively a character-level problem: the decision belongs to a single vowel, at which a subword tokenizer offers no position. It is also the prerequisite for \emph{metrical scansion}, since a syllable's weight depends on the quantity of its nucleus wherever the coda leaves it undetermined. The third experiment therefore tests the shared design of \S\ref{sec:design} on a task where character-level resolution is not merely convenient but constitutive. We evaluate both on \emph{Norma Syllabarum Graecarum} (Norma), the hand-annotated benchmark of \citet{thorncleland2026vowel}.

\paragraph{Model and training data.}
Two linear heads sit on the Stoicheia backbone and are trained jointly: a macron head (long versus short at ambiguous bare $\alpha$/$\iota$/$\upsilon$) and a per-letter scansion head (none, heavy-end, light-end, verse-end). At inference an optional Viterbi decoder constrains the scansion output to valid paths through a set of metre automata. The training labels are silver throughout. 177K verse lines carry vowel lengths: 59K converted from Hypotactic's syllable-weight markup \citep{hypotactic} wherever an open, coda-less syllable's weight determines its nucleus, and 118K mined by a constraint solver that accepts a line only if exactly one metrical grammar scans it and fixes a dichronon only where every admissible parse agrees. A further 1.6M lines of OGA prose, macronized by the rule-based system of \citet{thorncleland2026vowel}, are sampled 150K per epoch -- the same silver the transformer baseline of Table~\ref{tab:macronizer} learns from. The scansion head trains on 66K bracketed verses split by work. Every source was screened against Norma and Hypotactic to prevent leakage. As in \S\ref{sec:morphosyntax}, the backbone has read editions of the poems Norma annotates: what is held out is the annotation, not the text.

\paragraph{Macronization.}
Table~\ref{tab:macronizer} compares all systems on Norma's 1{,}916 test positions. Coverage separates them as much as accuracy does: the rule-based macronizer of \citet{thorncleland2026vowel} abstains on 40\% of the positions, and even scoring every abstention as short -- the majority class, its most favourable treatment -- leaves it at 89.46. Their 0.87M-parameter character transformer, trained on that system's output, abstains on 6. Stoicheia never abstains and averages 93.37 over the six fine-tuning seeds of the ablation below (range 89.61--94.52), 2.8 points above that dedicated model.

\paragraph{Scansion.}
The jointly trained model's scansion head labels every letter and reaches 89.74 balanced accuracy on Norma (88.74 plain accuracy, 87.89 syllable-boundary F1). Adding the Viterbi decoder raises exact-line accuracy on a held-out work split from 77.87 to 78.45, positive on all six seeds; on the random-init arm it is worth +1.27, roughly twice as much. An external metrical constraint, in other words, substitutes for what a weaker model has not internalized, and its diminishing returns are themselves a measure of how much of the metre the pretrained backbone already encodes. No prior system reports syllable-weight scansion on this benchmark, so the random-init control is the only comparison available.

\paragraph{Pretraining ablation.}
Against its matched random-init control, six seeds per arm, pretraining is worth +6.0 points of balanced accuracy on macronization and +2.2 on scansion (Table~\ref{tab:meter}). Macronization is measured on the macron-only arm and scansion on the joint one, each against its own control; the joint model macronizes about a point worse, the price of a shared backbone. Only the macronization effect is significant, at the $n{=}6$ floor of the exact test ($p{=}0.031$, $d_z{=}15.9$), with all six differences positive; for scansion $p{=}0.0625$ and one difference is negative.

The asymmetry is this experiment's most informative result. Syllable weight is largely computable from visible orthography, so a randomly initialized model can learn most of scansion from the silver corpus alone. But the length of a dichronon is not recoverable from local orthographic evidence. It is lexical and morphological knowledge that must already be present before the fine-tuning data can be exploited, and the +6.0-point gap is the price of not having it.

\section{Conclusion}
\label{sec:conclusion}

The introduction posed four desiderata: open data, character-level resolution, separable editorial layers, and guaranteed ignorance. Our artifacts meet them: an open, revision-pinned corpus (\S\ref{sec:data}); Stoicheia, whose five character planes are independently maskable (\S\ref{sec:model}); and eleven checkpoints such that for any passage at least one has never read it. Three experiments, each with a matched random-init control, test what this contributes. On restoration -- the one task the DeepMind systems and ours share -- the documentary-clean model beats Ithaca and its Aeneas-framework successor on Ithaca's own split, and both, plus an 8B instruction-tuned LLM, on documents edited after every compared system's cutoff (\S\ref{sec:restoration}); it exceeds the tuned subword baselines and every published figure on Celano's treebank (\S\ref{sec:morphosyntax}); and it beats the rule-based and neural macronizers on Norma while scanning verse from the same weights (\S\ref{sec:meter}). In each case the control shows the margin comes from pretraining rather than capacity or recipe. For any inscription or papyrus, an editor can now choose a model that never read it, and so read the tradition with an eye that cannot be recalling it.

\section{Limitations}
\label{sec:limitations}

\paragraph{Most of the pretraining corpus is machine-repaired.}
The repaired tier ($\sim$271M of $\sim$361M words) was reconstructed by an instruction-tuned LLM constrained to correct only well-attested OCR confusions and to discard rather than fabricate. We report no manual audit of how faithfully that constraint was observed, and the gap between the model's dev bits-per-character on pristine and repaired reference slices (0.234 against 0.309) indicates that the tiers are not interchangeable in quality. Downstream fine-tuning data is unaffected, but pretraining rests in large part on text no human has verified.

\paragraph{The decontamination guarantee is about strings, not about knowledge.}
Our fold exclusion matches exact and reordered duplicates, word 5-grams, and document-level near-duplicates, so a released model provably never saw a given passage \emph{as text}. It may still have read a scholion, commentary, lexicon entry, or later paraphrase. The guarantee we can make is verbatim non-exposure.

\paragraph{Artificial gaps are not real damage.}
Lacunae in our protocol are sampled from surviving text, uniformly over $L{=}1$--10 characters, following \citet{assael2022ithaca}. Every compared system faces the same distribution, so the rankings are unaffected, but the absolute error rates should not be read as expected performance on an arbitrary damaged document. The protocol further assumes, as this line of work generally does, that the extent of a lacuna is known in advance from physical evidence; unknown-length gaps are not evaluated here, although the elastic masking of \S\ref{sec:model} trains the capability.

\paragraph{We always answer, and we do not say how sure we are.}
Stoicheia abstains on no macronization position, and we report top-1 and top-20 accuracy but no calibration or selective-prediction analysis. The rule-based macronizer's 40\% abstention rate is a designed property: it declines where the evidence does not decide. An always-answering model is more useful than an abstaining one only if an editor can tell which predictions to trust, and we do not yet provide that.

\paragraph{Neither the planes nor the objective is ablated.}
No experiment trains a subword-tokenized model, or a model with diacritics folded into a flat character vocabulary, under the same objective and corpus. The cross-encoder comparison of \S\ref{sec:morphosyntax} holds the downstream recipe fixed but varies pretraining objective, corpus, and input factorization jointly, so it establishes that the backbone is better without isolating which of its design choices makes it so. Relatedly, holding one recipe fixed across encoder families is what makes that comparison interpretable, but a learning-rate sweep is the only per-encoder adaptation we perform; the results should be read as ``this recipe, applied uniformly.''

\paragraph{Benchmark provenance.}
\emph{Norma Syllabarum Graecarum} and both macronization baselines of \S\ref{sec:meter} come from a single prior project \citep{thorncleland2026vowel}, so our evaluation on that task depends on one group's benchmark, annotation decisions, and system design. To avoid advantaging ourselves in scoring, the external systems are evaluated with that project's own script rather than ours, and the constant-short baseline -- which computes identically in both harnesses -- is reported so the two sets of figures can be checked against a common reference point.
\section*{Ethical Considerations}

\paragraph{Proposed readings can acquire unearned authority.}
A model that supplies plausible Greek for a lacuna produces text that looks like an edition. The danger is not that it is wrong but that it is fluent: a supplement adopted without independent argument can enter the scholarly record and be cited as though it rested on evidence. However, conjectural criticism has been a core practice of systematic philology since its inception in the 3rd century BC. The history of forgeries dates back to antiquity, and the scholarly community is
well versed in addressing the challenges posed by falsaria and the risks of overreliance on speculative textual scholarship.

\paragraph{Most of the corpus is machine-repaired.}
Roughly 271M of $\sim$361M words passed through LLM-based OCR correction. The tiers are released separately and labelled, and the pristine tier is untouched, so a user can choose the unrepaired subset; but anyone reusing the repaired tier as a philological source rather than as training data risks propagating uncorrected OCR errors into scholarship. The paper's own use is as pretraining material only.

\paragraph{Compute.}
Each of the eleven backbones was trained for roughly 42 hours on 128 NVIDIA GH200 GPUs, about 5.4K GPU-hours per model and some 59K GPU-hours in total; the documentary, morphosyntactic and metrical fine-tunes cost about one GPU-hour each. Releasing all eleven checkpoints, rather than only the flagship, is intended to make that cost non-recurring for others.

\paragraph{Data.}
The corpora consist of ancient and medieval texts and their modern editions; they contain no personal data about living individuals, and no human subjects were involved in this work. Ancient sources describe slavery, sexual violence, and ethnic hostility as ordinary features of their world, and a model trained on them reproduces that distribution; the release is intended for philological analysis, where confronting such content is part of the object of study.

\paragraph{Use of AI assistants.}
An AI coding and writing assistant (Anthropic's Claude, models Fable~5 and Opus~5) was used throughout this project under author direction: it contributed to the implementation of the training, evaluation and release code, all of which is public, and to the drafting and revision of prose, whose argumentative content originates with the authors. The authors take full responsibility for all content.

\section*{Disclosure}
\emph{Norma Syllabarum Graecarum}, the rule-based macronizer and the character transformer of
\S\ref{sec:meter}, and the comparison release whose recently edited documents and per-record
predictions are reused in \S\ref{sec:restoration}, are prior work by the authors of this paper
\citep{thorncleland2026vowel,cullhed2024instruct}. The safeguards this motivates in scoring are
described under ``Benchmark provenance'' in \S\ref{sec:limitations}.

\section*{Acknowledgments}
Computational resources were provided by the National Academic Infrastructure for Supercomputing in
Sweden (NAISS), funded by the Swedish Research Council.

\bibliography{custom}

\end{document}